\documentclass{article}

\usepackage{microtype}
\usepackage{graphicx}
\usepackage{caption}
\usepackage{subcaption}
\usepackage{booktabs} 

\usepackage{hyperref}

\usepackage[accepted]{icml2025}

\usepackage{amsmath}
\usepackage{amssymb}
\usepackage{mathtools}
\usepackage{amsthm}
\usepackage{enumitem}

\usepackage[capitalize,noabbrev]{cleveref}

\theoremstyle{plain}
\newtheorem{theorem}{Theorem}[section]

\newtheorem{lemma}[theorem]{Lemma}

\theoremstyle{definition}

\newtheorem{assumption}[theorem]{Assumption}
\theoremstyle{remark}
\newtheorem{remark}[theorem]{Remark}
\usepackage{xspace}
\usepackage{natbib}
\usepackage{bm}
\usepackage{enumitem}
\theoremstyle{definition}

\newcommand{\cX}{\ensuremath{\mathcal{X}}}

\newcommand{\cR}{\ensuremath{\mathcal{R}}}
\newcommand{\R}{\ensuremath{\mathbb{R}}}

\newcommand{\cY}{\ensuremath{\mathcal{Y}}}
\renewcommand{\P}{\ensuremath{\mathbb{P}}}

\DeclareMathOperator{\E}{{\mathbb E}}

\DeclareMathOperator*{\argmin}{\textrm{argmin}}

\usepackage{eqparbox}

\newcommand{\KL}{\mathrm{KL}}

\newcommand{\pref}{\omega}

\newcommand{\pisft}{\pi_{\text{ref}}}
\newcommand{\piref}{\pi_{\textrm{ref}}}

\newcommand{\pihat}{\ensuremath{\widehat{\pi}}}

\newcommand{\pistar}{\ensuremath{\pi^\star}}

\newcommand{\base}{\mu}

\newcommand{\pidata}{D_y}
\newcommand{\Rscale}{\mathfrak{R}}
\newcommand{\CE}{\textrm{CE}}

\usepackage{tcolorbox}

\usepackage[textsize=tiny]{todonotes}

\icmltitlerunning{Offline Learning Preference-based RL}

\begin{document}

\twocolumn[
\icmltitle{Design Considerations in Offline Preference-based RL}



\icmlsetsymbol{equal}{*}

\begin{icmlauthorlist}
\icmlauthor{Alekh Agarwal}{yyy}
\icmlauthor{Christoph Dann}{yyy}
\icmlauthor{Teodor V. Marinov}{yyy}
\end{icmlauthorlist}

\icmlaffiliation{yyy}{Google Research}
\icmlcorrespondingauthor{Alekh Agarwal}{alekhagarwal@google.com}
\icmlcorrespondingauthor{Christoph Dann}{chrisdann@google.com}
\icmlcorrespondingauthor{Teodor V. Marinov}{tvmarinov@google.com}

\icmlkeywords{Machine Learning, ICML}

\vskip 0.3in
]



\printAffiliationsAndNotice{\icmlEqualContribution} 

\begin{abstract}
Offline algorithms for Reinforcement Learning from Human Preferences (RLHF), which use only a fixed dataset of sampled responses given an input, and preference feedback among these responses, have gained increasing prominence in the literature on aligning language models. In this paper, we study how the different design choices made in methods such as DPO, IPO, SLiC and many variants influence the quality of the learned policy, from a theoretical perspective. Our treatment yields insights into the choices of loss function, the policy which is used to normalize log-likelihoods, and also the role of the data sampling policy. Notably, our results do not rely on the standard reparameterization-style arguments used to motivate some of the algorithms in this family, which allows us to give a unified treatment to a broad class of methods. We also conduct a small empirical study to verify some of the theoretical findings on a standard summarization benchmark. 
\end{abstract}

\section{Introduction}
\label{sec:intro}

The now substantial literature on Reinforcement Learning with Human Preferences (RLHF) can be broadly categorized into two families of methods. Given a dataset of human preferences, the first class of online algorithms are based on learning a reward function that assigns numerical scores to a response $y$ given some input $x$, such that the high-scoring responses are preferred over the low-scoring ones in our preference dataset. These methods subsequently maximize this reward function using an \emph{online RL} algorithm like PPO~\citep{christiano2017deep, stiennon2020learning, ouyang2022training} or Reinforce~\citep{ahmadian2024back}. A different approach forgoes the reward learning step and uses a reparameterization trick to directly learn a good policy from the preference dataset, with Direct Preference Optimization (DPO)~\citep{rafailov2024direct} being the pioneering approach in this line of work. These approaches are referred to as \emph{offline or direct alignment}, since they do not draw any fresh samples from the learned policy during training, and only use the responses observed in the preference dataset. This paper focuses on this latter family of algorithms, and studies how the properties of the data and the learning objective affect the quality of the learned policy. 

Offline methods for RLHF such as DPO~\citep{rafailov2024direct}, IPO~\citep{azar2024general}, SLiC~\citep{zhao2022calibrating, zhao2023slic} and KTO~\citep{ethayarajh2024kto}, along with a growing number of variants have received a significant attention in the academic literature owing to a number of attractive properties. First, the removal of an explicit reward learning step simplifies the number of modeling choices and steps required for the RLHF pipeline, along with reducing the demand on computational resources. Furthermore, the requirement to only evaluate a policy's likelihood on a fixed set of responses in the preference dataset, as opposed to generating responses from the learned policy as in online RL, adds further resource efficiency. However, the growth of this literature has also spurred an equally large body of work now detailing the various deficiencies of these techniques, such as the tendency of the algorithms to shift the probability mass outside the support of the observed responses in the preference dataset, significant preference hacking behaviors and notorious collapses in the learning dynamics with continued training~\citep{pal2024smaug, park2024disentangling, rafailov2024scaling, fisch2024robust}. 

While some of the issues raised above have received theoretical treatment for specific approaches, the literature still lacks a comprehensive theoretical foundation underlying these offline RLHF techniques. Most of these techniques have been motivated by some variant of the original reparameterization argument in the DPO paper, but the argument hinges on unformalized assumptions regarding the coverage of the data with respect to the learned policy, and does not capture all the algorithmic variants which have subsequently been developed in the literature. 

\begin{table*}[t]
    \centering
    \begin{tabular}{c|c|c|c}
    Method & Base policy $\mu(y | x)$ & Loss $\ell$ & Constraint/Regularizer\\
    \hline
        DPO~\citep{rafailov2024direct}& $\piref(y | x)$ & $\ell(z) = - \log\left(\frac{1}{1 + \exp(-\beta z)}\right)$ & -\\
        IPO~\citep{azar2024general}& $\piref(y | x)$ & $\ell(z) = (z - \tau)^2$ & -\\
        Slic-HF~\citep{zhao2023slic} & $1$ & $\ell(z) = \max\{\tau - z, 0\}$ & $\CE(\piref, \pi)$\\
        GPO~\citep{tanggeneralized} & $\piref(y | x)$ & $\ell(z) = f(\beta z)$ & -\\
        CPO~\citep{xu2024contrastive}& 1 & $\ell(z) = - \log\left(\frac{1}{1 + \exp(-\beta z)}\right)$ & $\CE(\pi_w, \pi)$\\
        R-DPO~\citep{park2024disentangling}& $\piref(y | x)\exp\left(\frac{\alpha}{\beta}|y|\right)$ & $\ell(z) = - \log\left(\frac{1}{1 + \exp(-\beta z)}\right)$ & -\\
        ODPO\footnotemark~\citep{amini2024direct}& $\piref(y | x)$ & $\ell(z) = - \log\left(\frac{1}{1 + \exp(-\beta z + \tau)}\right)$ & -\\
        SimPO\footnotemark~\citep{meng2024simpo} & 1 & $\ell(z) = - \log\left(\frac{1}{1 + \exp(-\beta z)} - \gamma\right)$ & -\\
    \end{tabular}
    \caption{A collection of methods for offline RLHF from preference feedback, along with the instantiation of the different design choices that make them a special case of our general framework. For ODPO and SimPO, they are not included in the full generality in our framework as discussed in the footnotes.}
    \label{tab:existing_methods}
\end{table*}

In this paper, we instead adopt the perspective of offline RLHF as just solving a \emph{loss minimization problem} on a preference dataset, and study when the optimal solution of this loss minimization yields a desirable policy. A key challenge in undertaking such a study is the identification of a benchmark policy which is both desirable in terms of the responses it generates, and is attainable under reasonable assumptions using offine RLHF. With this background, our paper makes the following contributions.

\begin{enumerate}[leftmargin=10pt, parsep=2pt, itemsep=0pt, topsep=0pt, partopsep=0pt]
\item We identify a benchmark policy to measure the performance of offline RLHF against. Prior work~\citep{swamy2024minimaximalist} shows that the class of offline techniques considered here cannot attain the optimal policy for online RLHF in general, and we instead develop a weaker benchmark under reasonable assumptions on the data generating process and the learning setup. For DPO, this benchmark does correspond to the optimal softmax policy when the preference data follows the Bradley-Terry-Luce~\citep{bradley1952rank, luce2012individual} model.
\item We provide a bound on the sub-optimality of the learned policy to the yardstick identified above in a learning framework that encapsulates most existing offline RLHF variants. The bound depends on the curvature of the loss function, as well as the coverage of the offline dataset.
\item We corroborate some of our theoretical findings using an empirical study on a summarization task, and find that squared loss of IPO outperforms the logistic loss of DPO due to its nicer curvature properties, while dropping the normalization with reference policy likelihood causes a small, but consistent deterioration in the quality of the learned policy, as predicted by our theory.
\end{enumerate}

\section{Problem Setup}
\label{sec:setup}

\paragraph{Preference-based learning.}
We consider the setting of learning from offline preference data. In this setting, we are typically given a dataset of samples where each sample consists of an input $x\in\cX\sim D_{x}$, two responses $y, y'\in\cY\times\cY \stackrel{\text{i.i.d.}}{\sim} \pidata(\cdot | x)$ and a binary label $\pref \in \{-1,1\} \sim D_\pref(\cdot | x, y, y')$. We use the shorthand $(x, y, y', \pref) \sim D_{xy\pref}$ to succinctly denote samples from this generative process. The label $\pref$ captures our preference for the response $y$ over $y'$, for the input $x$. We study offline preference-based RL methods, which are parameterized by some loss function $\ell~:~[-\Rscale,\Rscale]\to [0, \infty)$. In particular, we study methods which take as input a policy class $\Pi\subseteq\{\cX\to \Delta(\cY)\}$, and find a policy which minimizes the following loss, given $n$ samples:
\begin{equation}
    \pihat = \argmin_{\pi\in\Pi} \sum_{i=1}^n\ell\biggr(-\pref_i\bigg(\underbrace{\log\frac{\pi(y_i|x_i)}{\base(y_i|x_i)} - \log\frac{\pi(y_i'|x_i)}{\base(y_i'|x_i)}}_{:=\pref_i^{\pi, \mu}}\bigg)\biggr).
    \label{eq:pref-loss}
\end{equation}
Here $\base \colon \cX \rightarrow \R_+^{\cY}$ refers to an arbitrary base policy, which does not need to be normalized. Typically $\ell$ is chosen so that it incentivizes $\pref_i\pref_i^{\pihat, \mu}$ to be positive. This means that $\pihat$ tends to increase the probability of producing $y_i$ compared with $y_i'$, relative to the base probabilities under $\base$, when $\pref_i = 1$. In the particularly simple case when $\base$ is just the uniform policy, we see that $\pihat$ results in a larger probability of producing $y_i$ over $y_i'$, when $\pref_i = 1$, which captures the basic intuition behind offline preference-based RL.
\addtocounter{footnote}{-1}
\footnotetext{ODPO in the general case allows a margin of the form $f(\text{score}(x,y) - \text{score}(x,y'))$, which is not admissible in our setup as specified, though it can be incorporated in our analysis with some additional work. When $f$ is the identity function, we can simply define $\base(y|x) \propto \piref(y|x) \text{score}(x,y)$, as a special case.}
\addtocounter{footnote}{1}
\footnotetext{SimPO in the general case has an additional length normalization, leading to the loss $\log(\pref\text{sigmoid}(\beta\log\pi(y|x)/|y| - \beta\log\pi(y'|x)/|y'|) - \gamma)$. This length normalization is currently not included in our theoretical setup and analysis.}
\paragraph{Choice of policy class.} A common policy parameterization is to use softmax policies $\pi_\theta(y | x) \propto \exp(f_\theta(y|x))$ with some parameter set $\theta\in\Theta$, where $f$ is some fixed network architecture. While we could explicitly define the policy class $\Pi$ this way, we intentionally leave $\Pi$ general at this point, to allow our framework to further include:
\begin{itemize}
    \item Additional constraints~\eqref{eq:pref-loss}: 
    Several works (e.g. \citet{zhao2023slic, xu2024contrastive}) regularize or constrain the preference objective of Equation~\ref{eq:pref-loss} with a cross-entropy term based on $\CE(\pi_0, \pi) := -\E_{x\sim D_x, y\sim \pi_0(\cdot | x)} \ln \pi(y|x)$, where $\pi_0$ is some policy of a reasonable quality. This is done by adding $\alpha \CE(\pi_0, \pi)$ to the objective or a constraint that $\CE(\pi_0, \pi) \leq \lambda$. The addition keeps the loss optimization from degenerating when the distribution $D$ underlying our data has a limited support, and the optimal policy $\pihat$ might produce responses outside the support of $D_{xy}$, where we do not have preference feedback. We capture such additional constraints or regularizers as optimization with a reduced policy class $\Pi$.
    \item Early stopping:  
    A different way to constrain the extent of optimization is directly in the parameter space. Most often the preferred optimizer is a first-order method such as Gradient Descent, AdaGrad, Adam or AdaFactor. Early stopping optimization with first-order methods directly corresponds to a constraint in the parameter space $\Theta$~\citep{yao2007early, raskutti2014early, neu2018iterate, suggala2018connecting, vaskevicius2019implicit, sonthalia2024regularization}.
For example, if GD is used as the optimizer, early stopping would correspond to an $\ell_2$ distance bound in parameter space between the parameters underlying $\pi$ and the initial policy $\pi_0$ at the start of optimization.
Early-stopping or limited training in typical fine-tuning setups is a common practice that we again capture through an appropriate choice for $\Pi$, e.g. with additional $\ell_2$ constraints on the policy parameters.
\end{itemize}

We assume for simplicity that the policy class $\Pi$ and the base policy $\base$ are chosen such that $\log\frac{\pi(y|x)}{\base(y|x)} - \log\frac{\pi(y'|x)}{\base(y'|x)} \in [-\Rscale, \Rscale]$ for $x, y, y'$ drawn according to the generative process described above, with probability 1. Consequently, the loss on each sample is also bounded by some $B$, and we expect that the empirical loss minimizer $\pihat$ from \eqref{eq:pref-loss} has also small population loss:
\begin{align}
    L_\base(\pihat; D_{xy\pref}) - \min_{\pi\in\Pi} & L_\base(\pi; D_{xy\pref} ) \leq \epsilon_n, \quad \textrm{with}
    \label{eq:loss-dist}\\
    L_\base(\pi; D_{xy\pref}) =&\, \E_{(x, y, y', \pref) \sim D_{xy\pref}}[\ell_\base(\pi; \pref, x, y, y')],\nonumber\\
    \ell_\base(\pi; \pref, x, y, y') =&\, \ell(\pref \cdot \pref^{\pi, \mu}(x, y, y')).\nonumber
\end{align}
The error bound $\epsilon_n$ typically scales as $O(\sqrt{Bd_\pi/n})$, with $d_\pi$ being the statistical complexity of $\pi$, such as $\ln |\Pi|$ for finite classes, log-covering number for the infinite case, or other complexity measures like Rademacher or Gaussian complexities. We abstract such treatment into a general error term $\epsilon_n$, as this analysis of the empirical loss minimizer is standard and not key focus of our study.

\paragraph{Existing offline RLHF methods.} The formalization of preference-based learning above captures a wide range of existing offline RLHF methods through appropriate choices of the base policy $\mu$ and loss $\ell$. Table~\ref{tab:existing_methods} presents a selection of methods that fit the setup and which our formulation captures. We note that the methods predominantly vary along three axes: the loss function $\ell$, the choice of the base policy $\base$ and the choice of the constraint or regularizer to guide the policy optimization. In the next section, we present a theoretical framework and our main technical results on the quality of the learned policy $\pihat$, with an emphasis on understanding the effects of these choices. We note here that the GPO paper of~\citet{tanggeneralized} considers an almost identical set of design choices (other than the flexibility in $\base$ and the choice of $\Pi$) as our work, but their emphasis is on empirical evaluation while we seek to understand the design space in theory.

\section{Analysis Framework and Main Results}
In this section, we set up a framework for analyzing offline preference learning algorithms which optimize~\eqref{eq:pref-loss}, and present our main results. We begin with a discussion to set up the performance criterion. 

\subsection{Analysis Framework}
\label{sec:framework}
\paragraph{Performance criterion.} How to measure the efficacy of a preference-based learning technique? As described above, based on our choices of $\ell$, $\base$ and $\Pi$, we get a guarantee on the expected loss of the resulting policy. But we want to measure how well the policy $\pihat$ does in terms of producing highly preferred outputs $y$, given inputs $x$. It is not clear that a policy which has a small loss also produces good outputs. For instance, suppose that the learned $\pihat$ is such that $\E[\pref | x, y, y'] \pref^{\pihat, \base}(x, y, y') > 0$ for any $x, y, y'$ in the support of our training distribution, and the base policy $\base$ is uniform. In this case, we can conclude that $\pihat(y | x) > \pihat(y' | x)$ whenever $\E[\pref | x, y, y'] > 0$. But this does not preclude the two probabilities from being extremely close, though correctly ordered, and more generally $\pihat$ might still place a non-trivial probability on the least desirable outputs $y$ for many inputs $x$. 

Ideally, we would like to say that $\pihat$ places most of the mass on the most desirable outputs $y$. Since the conditional probabilities $D_\pref(\cdot | x, y, y')$ can be interpreted as a preference function $\P(y \succ y' | x)$, one notion of an optimal policy is provided by the Nash equilibrium policy for the two player-game encoded by this preference function, as considered in prior works~\citep{wang2023rlhf, munos2023nash, swamy2024minimaximalist}. However, as \citet{swamy2024minimaximalist} show, this optimal solution cannot be attained by minimizers of the objective~\eqref{eq:pref-loss} in general, where they provide a lower bound for the special case of DPO. Consequently, we need a different yardstick to measure our performance for the setup of offline preference-based RL, which we do next. We begin with introducing some useful notation and our formal assumptions needed to define the benchmark policy. 

Given the form of the loss in Equation~\ref{eq:pref-loss}, we have to reason about the log probabilities as the main object of interest. It is therefore convenient to denote for each policy $\pi$ its log probabilities by $R_{\pi}$ with $R(x, y) = \log \pi(y | x)$, and conversely by $\pi_R$ the policy associated with $R$. 
We will also refer to such $R$ as a \emph{reward function} since it measures the quality of the output $y$ for input $x$, and the policy $\pi_R$ ascribes higher probability to outputs with high rewards under $R$.
For the policy class $\Pi$ we can then define the accompanying reward class as $\cR = \{ R_{\pi} \colon \pi \in \Pi \}$. Note that this is only nomenclature and is not a modeling assumption such as a reward-based Bradley-Terry-Luce (BTL) model of preferences in the data generating process.

\paragraph{Modeling assumptions.}
We start by assuming that the log-probabilities of all policies and the base policy $\mu$ is bounded, which ensures that the inputs of the loss function are in a bounded range. Additionally, we assume that also the loss outputs are bounded which holds for all practical loss functions, given the bounded domain $[-\Rscale, +\Rscale]$.
\begin{assumption}
    \label{ass:uniform_bound}
    For all $x,y,y'$ and all $\pi\in\Pi$, we have  $|R_\pi(x,y)| \leq \frac{\Rscale}{4}$, $|\log \mu(y | x)| \leq \frac{\Rscale}{4}$ and $\ell_\base(\pi; \pref, x, y, y') \leq B$.
\end{assumption}

Given the policy class $\Pi_\cR$, we now make a realizability assumption on the data generating mechanism with respect to this class.

\begin{assumption}[Realizability]
    There exists $\pistar \in \Pi$ such that for all $ x \in \cX, y \in \cY, y' \in \cY$:
    \begin{align*}
    \pref^{\pi^\star,\base}(x,y,y') = \argmin_{v \in [-\Rscale, \Rscale]}\E_{D_\pref}[\ell(\pref\cdot v) | x, y, y']
    \end{align*}
\label{ass:realizable}
\end{assumption}
A necessary condition for Assumption~\ref{ass:realizable} to hold is that there is a fixed policy $\pi^\star$ which minimizes the loss $\ell_{\base}$ in a pointwise manner for all $x, y, y'$, when we take conditional expectation only over the preference labels $\pref$. The assumption further requires that within the range of $R_\pi$ which parameterizes $\pi \in \Pi$ we have that $\pi^\star$ is the pointwise minimizer of $\ell(\pref\cdot v)$. This makes the optimal policy $\pi^\star$ independent of the distributions $D$ and $\pidata$ over $x, y, y'$. To further understand why it is helpful to have such an optimal policy $\pi^\star$, we make a standard calibration assumption on the loss function $\ell$ in Equation~\ref{eq:pref-loss}.

\begin{assumption}[Proper loss]
    We assume that the loss function $\ell$ is a proper loss for class probability estimation~\citep{reid2010composite}. That is, there is a function $g_{\ell}$ which depends only on $\ell$, such that for all $\eta \in [0,1]$:
    $\argmin_v \eta \ell(v) + (1-\eta) \ell(-v) = g_\ell(\eta)$.
    \label{ass:proper}
\end{assumption}
That is, when we take conditional expectation over the binary label according to probability $\eta$, then the minimizer of the loss correctly recovers some loss-dependent function of $\eta$. This condition is satisfied by most commonly used differentiable losses for binary classification such as the logistic loss, squared loss, squared hinge loss etc. For instance, the function $g_{\ell}$ is given by $g_{\text{sq}}(\eta) = 2\eta-1$ for squared loss and $g_{\text{log}}(\eta) = \ln(\eta/(1-\eta))$ for the logistic loss.

\paragraph{Realizability, proper losses, and optimal policies.} When we use proper losses, the realizability condition takes a particularly intuitive form when the data generating process and the $g_\ell$ function underlying the loss $\ell$ agree with each other. For instance, suppose we use the logistic loss and the preferences are generated according to a BTL model: $P(\pref = 1 | x, y, y') = 1 / (1 + \exp(R^\star(x,y') - R^\star(x,y)))$ for some $R^\star \in \cR$. Then under Assumptions~\ref{ass:realizable} and~\ref{ass:proper}, we have that $\pi^\star = \pi_R$ for $R$ such that for any $x, y, y'$
\begin{equation}
\begin{aligned}
    R(x, y) - R(x, y') &= R^\star(x, y) - R^\star(x, y')\\
    &+ \ln \base(y|x) - \ln \base(y' | x).
    \label{eq:R-realizable}
\end{aligned}
\end{equation}
That is, $R$ is given by $R^\star + \ln \base$, up to an $x$ dependent offset, within the support of the data. A similar conclusion holds for the squared loss, and if $P(\pref = 1 | x, y, y') = 0.5 + (R^\star(x, y) - R^\star(x, y'))/2$.
A more detailed discussion of how modeling $D_\pref$ as part of the exponential family leads to proper losses, $\ell$, can be found in Appendix~\ref{app:preference_to_loss}.
We see that in these cases, the policy underlying realizability learns a \emph{ground-truth} reward function which underlies our data generation process. In such a scenario, where there is a reward function $R^\star$ underlying the observed preferences, a natural benchmark is the KL-regularized reward maximizing policy, $\pi(y|x) \propto \pi_0(y|x)\exp(\beta R^\star(x,y))$, where $\pi_0$ is some base policy with (such as the SFT policy), with respect to which the KL divergence is defined. When $\base = \pi_0$, then we see that the policy $\pi^\star$ exactly corresponds to this optimal policy for an appropriately chosen loss function. 

\paragraph{Performance criterion under realizability.} Based on these insights, we adopt the policy $\pi^\star$ as our performance yardstick, and seek a policy $\pi$ to minimize $\KL(\pi^\star||\pi)$. Approximately minimizing $L_\base$ may not be sufficient to derive meaningful bounds on $\KL(\pi^\star||\pi)$, however, as there still needs to be alignment between the data-generating distribution $D_{y}$ and $\pi^\star$. Otherwise, if there is no good coverage of the support of $\pi^\star(\cdot|x)$ by $D_{y}(\cdot|x)$ there is no guarantee that $\pi$ will be able to distinguish good responses, $y$, according to $\pi^\star$ from highly sub-optimal ones. This necessitates making the following coverage assumption.
\begin{assumption}[Coverage for optimal policy]
    \label{ass:covarage}
    Let $\bar R(x,y) = R(x,y) - \E_{D_y}[R(x,y)|x]$ denote the $y$-centered reward and let $R^\star$ denote the parameterization of $\pi^\star$. Further, let $\Delta \bar R(x,y) = \bar R(x,y) - \bar R^\star(x,y)$. We assume that there exists a constant $C$ s.t. for $R\in \cR$ it holds that
    \begin{align*}
        \E_{x,y \sim \pi^\star(\cdot|x)}[\Delta \bar R(x,y)^2] \leq C \E_{x,y\sim D}[\Delta \bar R(x,y)^2].
    \end{align*}
\end{assumption}
The coverage condition is akin to generalized coverage conditions used in the offline RL literature~\citep{xie2021bellman, jiang2024offline}. A sufficient condition to ensure this holds is to have $\sup_{x,y}\frac{\pi^\star(y | x)}{D_y(y | x)} \leq C$, but the generalized notion also holds when the class $\cR = \{w^\top\phi(x,y)~:~\|w\|_2\leq 1\}$ and we have that $\lambda_{\max}(\Sigma_y^{-1/2}\Sigma_{\pi^\star_\mu}\Sigma_y^{1/2}) \leq C$. Here we denote $\Sigma_\pi = \E_{(x\sim D, y\sim \pi(\cdot | x)}[\phi(x,y)\phi(x,y)^\top]$, and abbreviate $\Sigma_y = \Sigma_{D_y}$. Clearly, this second condition can be much weaker than the density ratio assumption, and indeed has underpinned several methods that effectively handle coverage issues in offline RL with large function spaces and high-dimensional data, motivating our definition here.

Before stating our main result we need a somewhat standard curvature assumption on the loss.
\begin{assumption}[Curvature around optimum]
    For any policy $\pi\in\Pi$ and functions $R, R^\star \in\cR$ such that $\pi = \pi_R, \pi^\star = \pi_{R^\star}$, there is a constant $c_\base > 0$ such that 
    \begin{align*}
     L_\mu(\pi; D_{xy\pref}) -& L_\mu(\pi^\star_\mu; D_{xy\pref})
    \geq\E\big[\pref\ell'(\pref \pref^{\pi^\star, \mu}(x, y, y'))\\
    \cdot&(\pref^{\pi, \mu}(x, y, y') - \pref^{\pi^\star, \mu}(x, y, y'))\big]\\
    +& \frac{c_\base}{2}\E\left[\big(\pref^{\pi^\star, \mu}(x, y, y') - \pref^{\pi, \mu}(x, y, y')\big)^2\right]
    \end{align*}
    \label{ass:curvature}
\end{assumption}
A sufficient condition for Assumption~\ref{ass:curvature} is to instead have the stronger condition that for any $u, v \in [-\Rscale, \Rscale]$, we have 
\begin{align*}
    \ell(u) \geq \ell(v) + \ell'(v)\cdot (u-v) + \frac{c_\base}{2} (u-v)^2.
\end{align*}
This condition holds for $c_\base = \frac{1}{2}$ for the squared loss: $\ell(u, v) = (u-v)^2$, and with an $\Rscale$-dependent constant for many other losses that are induced by log-likelihoods of exponential families, which includes the logistic loss and the probit loss. Assumption~\ref{ass:curvature} weakens this condition by requiring curvature on the expected loss $L$ only around the optimal policy $\pi^\star$, rather than pointwise on $\ell$.

\subsection{Main Results}

With our main modeling assumptions set up, we now give the main theoretical result on the KL divergence between an approximate minimizer of the population loss $L_\base(\pi; D_{xy\pref})$ and the benchmark $\pi^\star$.

\begin{theorem}
\label{thm:kl_bound}
For any $\pi \in \Pi$ such that $L_\base(\pi; D_{xy\pref}) - L_\base(\pi^\star_\mu; D_{xy\pref}) \leq \epsilon$, where the corresponding loss to $L_\base$, given by $\ell$ is proper, and under Assumptions~\ref{ass:uniform_bound}-\ref{ass:curvature}, it holds that
\begin{align*}
    \E_x\left[\KL(\pi^\star(\cdot | x)||\pi(\cdot | x))\right] \leq \sqrt{\frac{\epsilon}{c_\mu}} + \sqrt{\frac{C\epsilon}{c_\mu}} + \frac{e^{\Rscale}}{2}\cdot \frac{C\epsilon}{c_\mu}.
\end{align*}
\end{theorem}

\begin{remark}[Choice of loss function] Our bound scales inversely with the curvature constant, meaning that losses with high curvature will lead to more favourable bounds in Theorem~\ref{thm:kl_bound}.
The squared loss satisfies Assumption~\ref{ass:curvature} with $c_\base = \frac{1}{2}$ for any range of $\pref^{\pi^\star,\base}$ as it is strongly convex, while the squared hinge loss only satisfies Assumption~\ref{ass:curvature} in $(-\infty, 1)$, finally the logistic loss satisfies Assumption~\ref{ass:curvature} with a range dependent $c_\base$ as the loss becomes less curved as the $\pref^{\pi^\star,\base}$ approaches $\infty$. Our theorem suggests that optimizing the squared loss is ideal in terms of the final bound as $c_\base$ is constant and bounded away from $0$ across the full range of the loss. Squared loss is also a proper loss, so that the main assumption which might fail is realizability. We do note that the probabilistic model corresponding to squared loss naturally is less realistic than say, the BTL model, corresponding to the logistic loss. Nevertheless, the benefits of squared loss are verified by our experiments in Section~\ref{sec:experiments}. We note that a related discussion of the curvature properties comparing the squared and logistic cases can also be found in~\citet{azar2024general}.
\end{remark}

\begin{remark}[Choice of base policy] While the choice of base policy $\base$ does not appear directly in the KL bound, it influences the realizability assumption. As already discussed, under a proper loss such as the logistic loss and a corresponding reward model such as BTL, realizability becomes equivalent to having the reward model plus a $\log \base$ term be part of the reward space $\cR$. Further, the choice of $\base$ can change $\pi^\star$, and as we discuss in Section~\ref{sec:framework}, the choice of $\base = \pisft$ naturally yields a desirable $\pi^\star$. In our experiments we use two commonly studied choices of $\base$, the uniform policy which puts equal probability on all responses, and a SFT policy $\pisft$.
\end{remark}

\begin{remark}[Effect of constraints] Recall that we constrain the optimization problem to a policy class $\Pi \subseteq \Pi_{\cR}$ which captures any constraints that we incorporate such as CE to a SFT policy, $\pi_0$, or the implicit constraints induced by the choice of optimizer and early stopping. We note that the choice of $\Pi$ determines if the benchmark, $\pi^\star$, which satisfies the realizability and curvature assumptions has to be part of $\Pi$. In this context, a cross-entropy regularization to $\pisft$ essentially makes an assumption that $\pi^\star$ lies in the vicinity of $\pisft$. Biasing the reference policy in cross-entropy towards preferred responses such as in CPO~\citep{xu2024contrastive} can be further beneficial in ensuring the feasibility of $\pi^\star$.
\end{remark}

\begin{remark}[Connections with prior results]
As mentioned earlier, there is now a substantial literature on the degeneracies of DPO in particular, due to its popularity, with primarily empirical~\citep{park2024disentangling, rafailov2024scaling, fisch2024robust}, but also some theoretical results in these works that demonstrate that DPO tends to shift mass away from the support of the preference data, with probabilities of both preferred and dispreferred responses in the data rapidly degrading to zero. The reader might wonder how to reconcile these negative observations with our positive result on the loss minimizers of DPO-style losses. However, there are a few caveats which apply to the specific case of DPO. First, as we remarked earlier, the curvature constant $c_\base$ for DPO degrades exponentially fast with $\Rscale$. Further, since DPO does not control the log-likelihood ratios through regularization terms, the quantity $\Rscale$ can rapidly grow large empirically, as pointed out in multiple papers, and as we corroborate in the next section. In fact, some works on incorporating pessimistic reasoning in DPO~\citep{fisch2024robust, liu2024provably, cen2024value, huang2024correcting} result in adding regularization terms which partly mitigate some of these degeneracies.
\end{remark}

\section{Analysis}

The high-level reasoning to prove Theorem~\ref{thm:kl_bound} is the following. We first use Assumption~\ref{ass:curvature} to establish that any $\epsilon$-minimizer of $L(\pi_R; D_{xy\pref})$ admits a bound on the expected error in the centered rewards $\E_{(x, y)\sim D}[\Delta \bar R(x,y)^2]$. We then invoke the coverage condition of Assumption~\ref{ass:covarage} to translate this error bound to be under the benchmark policy $\pi^\star$. Subsequently, we relate the KL divergence between $\pi$ and $\pi^\star$ in terms of expectation of $\Delta \bar R(x,y)^2$ under $\pi^\star$, using a careful anaylsis of the log-partition function. 

We start with the first step in the sketch above.
\begin{lemma}
\label{lem:reward_bound}
Under Assumptions~\ref{ass:realizable} and~\ref{ass:curvature}, any policy $\pi\in\Pi_{\cR}$ with $L_\mu(\pi; D_{xy\pref}) - L_\mu(\pi^\star; D_{xy\pref}) \leq \epsilon$, satisfies
\begin{align*}
    \E_{x,y\sim D}&\left[\Delta \bar R(x,y)^2\right] \\&= \E_{x,y,y'\sim D}[\left(\Delta R(x,y) - \Delta R(x,y')\right)^2]\leq \frac{2\epsilon}{c_\base}.
\end{align*}
\end{lemma}
\begin{proof}
The first condition of the lemma is equivalent to
\begin{align*}
    &\E[\ell\left(\pref \pref^{\pi, \mu}(x, y, y')\right)] \geq \E[\ell\left(\pref \pref^{\pi^\star, \mu}(x, y, y')\right)]\\
    + &\E[\pref\ell'\big(\pref \pref^{\pi^\star, \mu}(x, y, y')\big)\big((\pref^{\pi, \mu}(x, y, y') - \pref^{\pi^\star, \mu}(x, y, y')\big)]\\
    +&\E[\frac{c_\mu}{2}\left(\pref(R(x,y) - R(x,y') - R^\star(x,y) + R^\star(x,y'))\right)^2]
\end{align*}
Now, Assumption~\ref{ass:realizable} implies that for any $\pi \in \Pi$,
\begin{align*}
    &\E_{\pref|x,y,y'}[\ell'\left(\pref \pref^{\pi^\star, \mu}(x, y, y')\right)\\
    \cdot&\left(\pref(\pref^{\pi, \mu}(x, y, y') - \pref^{\pi^\star, \mu}(x, y, y')\right)] \geq 0,
\end{align*}
where we have used the fact that $v^\star = \pref^{\pi^\star,\mu}(x,y,y')$ is the minimizer of $\E_{\pref|x,y,y'}[\ell(\pref\cdot v^\star)]$ together with first order optimality so that $\E_{\pref|x,y,y'}[\ell'(\pref v^\star)\pref(v - v^\star)] \geq 0$ for any $v$ and in particular for any, $\pi$, such that $v = \pref^{\pi,\mu}(x,y,y')$.
This, together with the second assumption of the lemma imply that
\begin{align*}
    \epsilon &\geq \E[\ell\left(\pref \pref^{\pi, \mu}(x, y, y')\right)] - \E[\ell\left(\pref \pref^{\pi^\star, \mu}(x, y, y')\right)]\\
    \geq &\frac{c_\mu}{2}\E[\left(\pref(R(x,y) - R(x,y') - R^\star(x,y) + R^\star(x,y'))\right)^2],
\end{align*}
where we have used the parametrization of $\pi$ and $\pi^\star$.
\end{proof}

Next we show how to bound the KL by $\Delta \bar R$ and the fraction of log-partition functions $\bar Z_\pi$ and $\bar Z_\star$, where $\bar Z_\pi(x) = \sum_{y} \exp(\bar R_\pi(x,y))$.
\begin{lemma}
\label{lem:kl_decomp}
Under Assumption~\ref{ass:covarage}, for any $\pi \in \Pi$ such that $\pi \propto \exp(\bar R(x,y))$, the expected KL divergence $\E_x\left[\KL(\pi^\star(\cdot | x)||\pi(\cdot | x))\right]$ is bounded by
    \begin{align*}
         \sqrt{2C\E_{x,y}[\Delta \bar R(x,y)^2]}
        + \E_x\left|\log\frac{\bar Z_\pi(x)}{\bar Z_\star(x)}\right|.
    \end{align*}
\end{lemma}
\begin{proof}
    \begin{align*}
        &\E_x[\KL(\pi^\star(\cdot | x)||\pi(\cdot | x))] \leq \E_x\left[\left|\E_{y\sim \pi^\star(\cdot | x)} \log\frac{\pi^\star(y | x)}{\pi(y|x)}\right|\right]\\
        \leq& \E_x\sqrt{2\E_{y\sim\pi^\star(\cdot|x)}\left(\log\frac{\pi^\star(y | x)}{\pi(y|x)} - \log \frac{\bar Z_{\star}(x)}{\bar Z_{\pi}(x)}\right)^2}\\
        +& \E_x\sqrt{2\E_{y\sim \pi^\star(\cdot | x)}\left(\log \bar Z_\pi(x) - \log \bar Z_\star(x)\right)^2}\\
        \leq& \sqrt{2C\E_{x,y}[\Delta \bar R(x,y)^2]} + \E_x \left|\log\frac{\bar Z_{\pi}(x)}{\bar Z_{\star}(x)}\right|,
    \end{align*}
where in second inequality follows from Jensen's inequality, and the third inequality follows from a combination of $(x+y)^2 \leq 2x^2+2y^2$ and $\sqrt{x+y} \leq \sqrt{x} + \sqrt{y}, \forall x,y\geq 0$, and the third inequality uses Jensen to push the expectation inside the square root, along with Assumption~\ref{ass:covarage}.
\end{proof}
The next lemma bounds the ratio of log-partitions.
\begin{lemma}
\label{lem:log_partition}
For any $\pi$ such that $L(\pi; D_{xy\pref}) - L(\pi^\star; D_{xy\pref}) \leq \epsilon$, it holds that
\begin{align*}
    \E_x \left|\log\frac{\bar Z_{\pi}(x)}{\bar Z_{\star}(x)}\right| \leq \sqrt{\frac{C\epsilon}{c_\mu}} + \frac{e^{\Rscale}}{2}\cdot \frac{C\epsilon}{c_\mu}
\end{align*}
\end{lemma}
\begin{proof}
Using the definition of $\bar Z_\pi$ we have
\begin{align*}
    \E_x \left|\log\frac{\bar Z_{\pi}(x)}{\bar Z_{\star}(x)}\right| &= \E_x \left|\log\sum_{y}\frac{\exp(\bar R_\pi(x,y))}{\bar Z_\star(x)}\right|\\
    &= \E_x \left|\log\sum_{y}\pi^\star(y|x)\frac{\exp(\bar R_\pi(x,y))}{\exp(\bar R_\star(x,y))}\right|.
\end{align*}
Here the second equation rearranges the definition $\pi^\star(y|x) = \exp(\bar R_\star(x,y))/\bar Z_\star(x)$. Proceeding further
\begin{align*}
    \E_x \left|\log\frac{\bar Z_{\pi}(x)}{\bar Z_{\star}(x)}\right| &=\E_x\left|\ln\E_{y\sim\pi^\star(\cdot|x)}\exp(\Delta \bar R (x,y))\right|\\
    &\leq \E_x\Bigg|\ln\bigg(1 + \E_{y\sim \pi_\star(\cdot |x)} \Delta \bar R(x,y)\\
    &+ \frac{e^{\Rscale}}{2} \E_{y\sim \pi^\star(\cdot |x)} \Delta \bar R(x,y)^2\bigg)\Bigg|\\
    &\leq  \E_x\bigg|\E_{y\sim \pi^\star(\cdot |x)} \Delta \bar R(x,y)\\
    &+ \frac{e^{\Rscale}}{2} \E_x\E_{y\sim \pi^\star(\cdot |x)} \Delta \bar R(x,y)^2\bigg|.
\end{align*}
Here the first inequality uses that $e^x \leq 1 + x + e^A x^2/2$, $\forall x \leq A$. The second inequality uses $\ln(1+x) \leq x$. Continuing with our simplification, we get the bound
\begin{align*}
    \E_x \left|\log\frac{\bar Z_{\pi}(x)}{\bar Z_{\star}(x)}\right|
    \leq& \sqrt{\E_x\E_{y\sim \pi^\star(\cdot |x)} \Delta\bar R(x,y)^2}\\
    &+ \frac{e^{\Rscale}}{2} \E_x\E_{y\sim \pi^\star(\cdot |x)} \Delta \bar R(x,y)^2\\
    &\leq \sqrt{\frac{C\epsilon}{c_\mu}} + \frac{e^{\Rscale}}{2}\cdot \frac{C\epsilon}{c_\mu}.
\end{align*}
The first inequality above follows by triangle inequality to the sum inside absolute value, followed by Jensen's inequality on the first term. Finally, we invoke Assumption~\ref{ass:covarage} and Lemma~\ref{lem:reward_bound} to control the last term, which completes the proof of the lemma.
\end{proof}
Combining Lemma~\ref{lem:kl_decomp} and Lemma~\ref{lem:log_partition} finishes the proof of Theorem~\ref{thm:kl_bound}.

\begin{figure*}[t!]
\hfill
\includegraphics[width=0.35\linewidth]{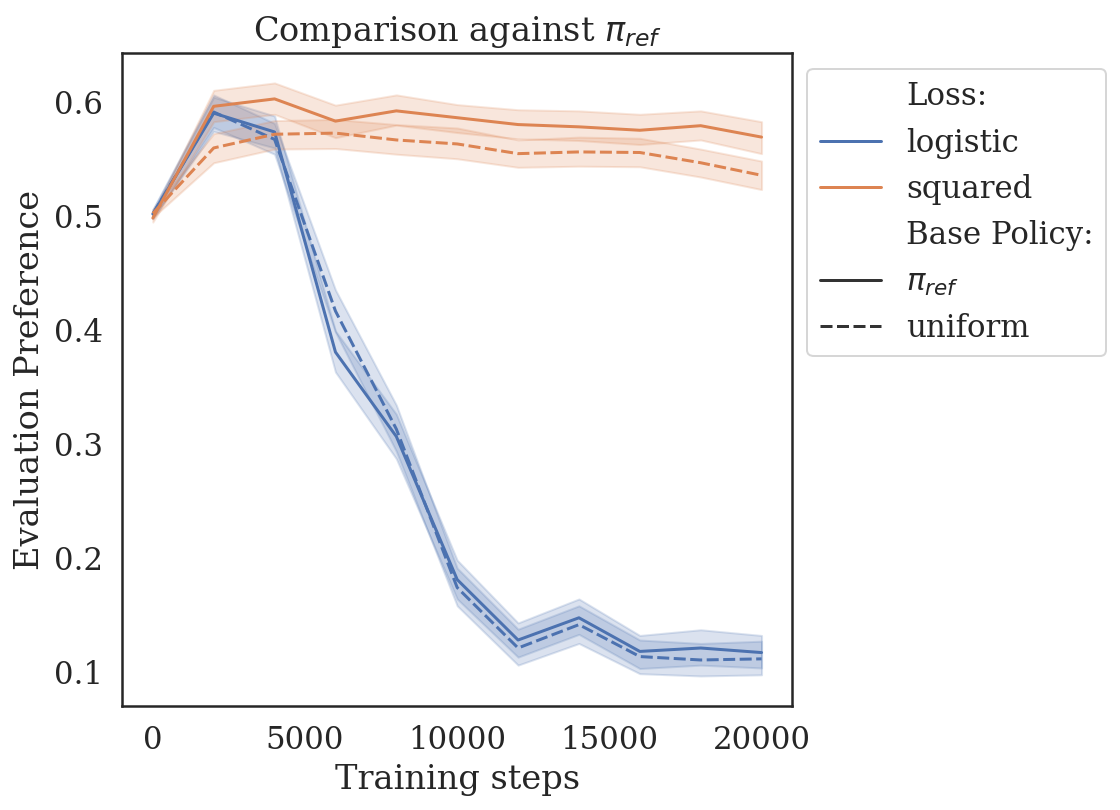}
\hfill
\includegraphics[width=0.35\linewidth]{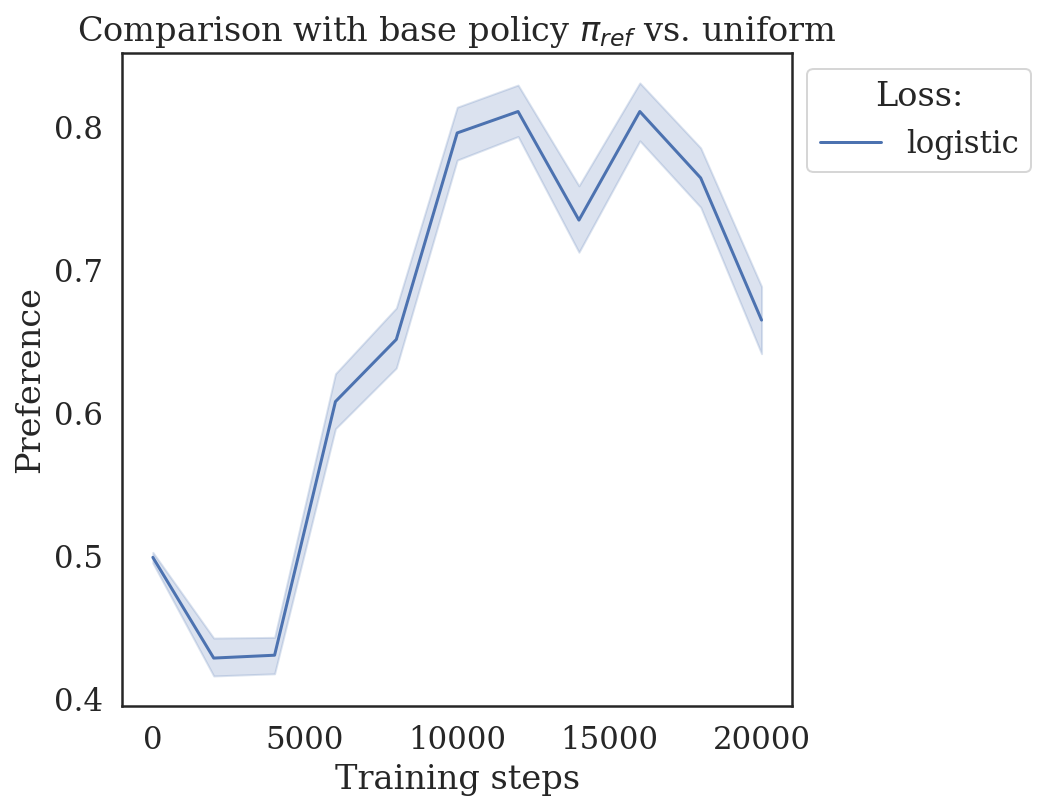}
\hfill
\caption{\footnotesize{\textbf{Left} panel shows the preference of the learned policy's summaries against those from the initial policy $\pisft$, as evaluated by a prompted Gemini 1.0 Ultra model. Shaded regions represent 95\% error bands. Both the logistic loss variants quickly improve in terms of the preference scores initially, but then suffer a catastrophic collapse. Squared loss improves at a similar rate initially, and remains stable throughout the training regime. \textbf{Right} panel shows a direct comparison between the variants of logistic loss using $\base = $ uniform and $\base = \pisft$ (DPO) at regular intervals in the training process. Interestingly, the uniform variant is preferred in the early stages of training, but as the training collapses around the training step 5K, the $\pisft$ variant starts to improve. Nevertheless, the absolute performance of both variants reaches its peak earlier in the training and rapidly worsens after 5K steps, suggesting that the preference for $\pisft$ over uniform in this region might not be particularly significant. See text for a more nuanced discussion.}}
\label{fig:eval}
\end{figure*}

\section{Experiments}
\label{sec:experiments}
We evaluate the impact of different design choices for offline RLHF methods on the standard TL;DR summarization task \citep{volske2017tl, stiennon2020learning}, where the task is to provide short summaries of articles. For our policy, we use a large T5 model~\citep{raffel2020exploring} with 770M parameters, which has been fine-tuned to maximize the log-likelihood of the human responses in the TL;DR dataset. This ensures that the policy is initialized with parameters that have reasonable likelihoods for the responses observed in our data. We experiment with two different losses, $\ell$:
\begin{align}
    \ell(x) &= \log(1 + \exp(-\beta x)) \tag{\text{Logistic loss}} \label{eqn:logistic}\\
    \ell(x) &= (\beta x - 1)^2, \tag{\text{Squared loss}}\label{eqn:squared}
\end{align}
where $\beta > 0$ is a hyper-parameter, governing the strength of how much each preference in the dataset should in the policy. 
We pair each of losses with two possible choices for the base policy $\mu$: The uniform base policy $\mu = 1$, which results in the following optimization
\begin{align*}
    \min_{\pi\in\Pi} \sum_{i=1}^n\ell\big(-\pref_i\big(\log \pi(y_i|x_i) - \log\pi(y_i'|x_i)\big)\big),
\end{align*}
and the second choice uses $\mu = \pisft$, the policy obtained after fine-tuning on TL;DR which our optimization is initialized with. This corresponds to the objective is:
\begin{align*}
    \min_{\pi\in\Pi} \sum_{i=1}^n\ell\big(-\pref_i\big(\log\tfrac{\pi(y_i|x_i)}{\pisft(y_i|x_i)} - \log\tfrac{\pi(y_i'|x_i)}{\pisft(y_i'|x_i)}\big)\big).
\end{align*}
We recall that when $\base = \pisft$, using the logistic loss from \eqref{eqn:logistic} corresponds to the DPO algorithm~\citep{rafailov2024direct} and the squared loss from \eqref{eqn:squared} corresponds to IPO~\citep{azar2024general}.

For each variant, we tuned the $\beta$ parameter in the loss in the interval $\{0.1, 0.5, 1.0\}$. For logistic loss, the best results are obtained at $\beta = 0.1$, while we did not see a significant difference across these choices for the squared loss, and show the results at $\beta = 0.5$. See Appendix~\ref{app:exp} for details.

\begin{figure*}[t!]
\hfill
\includegraphics[width=0.35\linewidth]{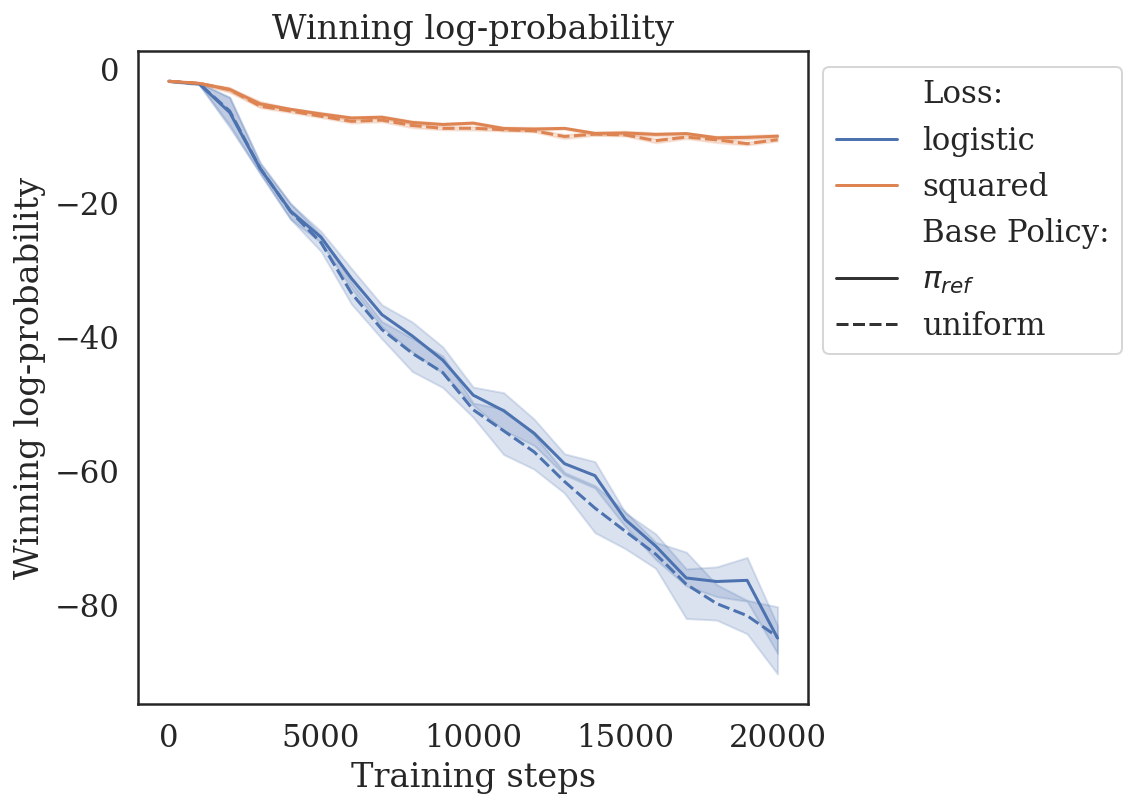}
\hfill
\includegraphics[width=0.35\linewidth]{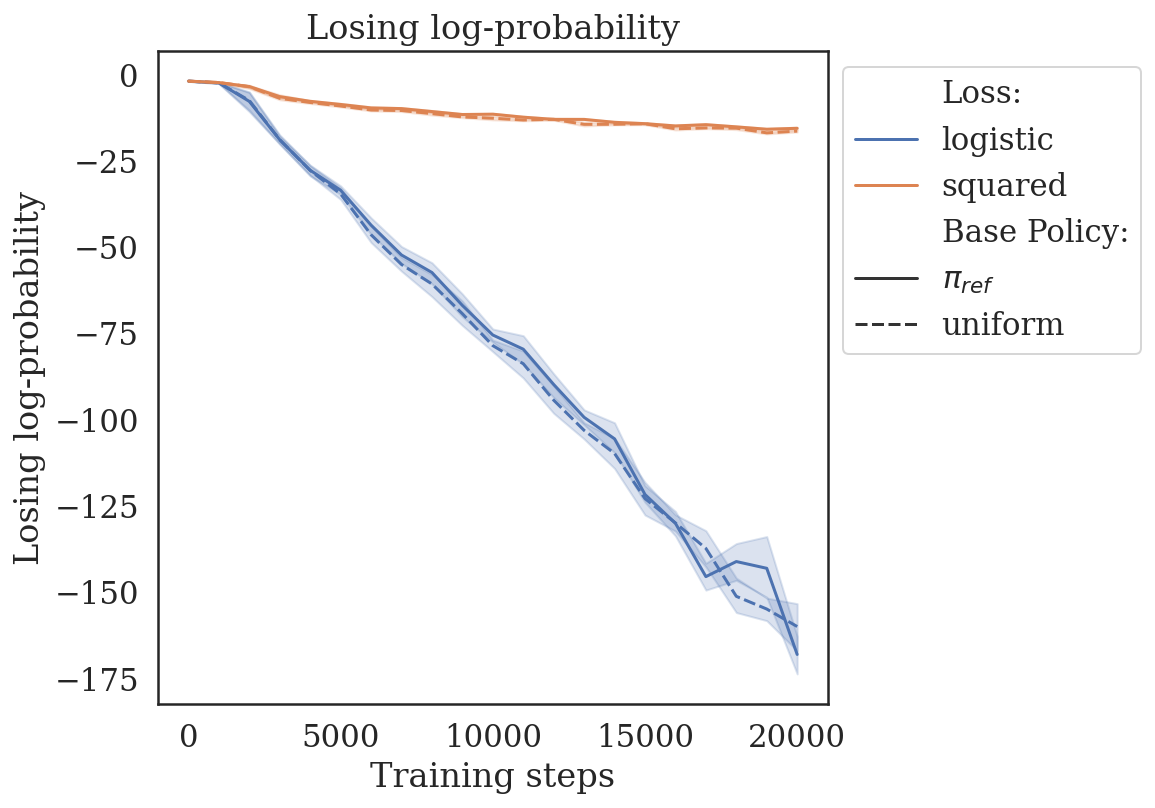}
\hfill
\caption{\footnotesize{Evolution of the log-likelihoods of the preferred response (\textbf{left}) and dispreferred response (\textbf{right}) from the preference dataset across the training process. Both variants of the squared loss decrease the log-likelihoods of both the responses during training, but the decrease is relatively mild. The logistic loss, on the other hand, sends these log-likelihoods crashing sharply, even though the dispreferred responses have significantly lower values, so the difference of log-likelihoods remains highly negative, driving the loss to zero. We suspect that this degeneration of log-likelihoods is responsible for the eventual collapse observed for the logistic loss in Figure~\ref{fig:eval}.}}
\label{fig:log_probs}
\end{figure*}

\begin{table*}[t]
    \centering
    \begin{tabular}{c|c|c|c}
    Loss & Base policy & $\Rscale$ & $c_\base$ \\
    \hline
        $\ell(z) = - \log\left(\frac{1}{1 + \exp(-\beta z)}\right)$ & $\mu = \piref$ & 73.203 & 0.00066\\
        $\ell(z) = - \log\left(\frac{1}{1 + \exp(-\beta z)}\right)$ & $\mu = 1$ & 72.949 & 0.00068\\
        $\ell(z) = (\beta z-1)^2$ & $\mu=\piref$ & 4.678 & 2.0\\
        $\ell(z) = (\beta z-1)^2$ & $\mu=1$ & 4.924 & 2.0\\
    \end{tabular}
    \caption{We approximate $\Rscale$ by $\omega^{\pi,\mu}$ averaged over the last mini-batch and approximate $c_\base$ by computing the curvature of $\ell_\mu$ at $\omega^{\pi,\mu}$.}
    \label{tab:c_base}
\end{table*}

We evaluate the different variants in two ways. First we compare the policies generated by each method at regular intervals in the training process, with the initial policy $\pisft$, by comparing the generated summaries in terms of their quality and conciseness by a prompted Gemini 1.0 Ultra model~\citep{team2023gemini}. 
Figure~\ref{fig:eval} (left) shows that squared loss variants perform significantly better than those with logistic loss, which reach a peak preference at 2k steps and suffer from a dramatic collapse afterwards. In the case of $\mu = \pisft$, this collapse  is consistent with prior findings on the DPO algorithm in the literature~\citep{fisch2024robust, rafailov2024scaling}. In comparison, both squared loss variants reach peak performance after similar number of steps, but maintain that performance more stably afterwards.

Notably, using $\mu = \pisft$ is consistently better than $\mu = 1$, when combined with the squared loss. In case of logistic loss, the comparison against $\pisft $ in Figure~\ref{fig:eval} (left) suggests that the base policy choice does not affect performance. However, when we compare the policies produced by the two variants with $\base = 1$ and $\base = \pisft$ directly in Figure~\ref{fig:eval} (left), we do observe a strong impact. Initially, the uniform variant is slightly preferred with a strong preference afterwards in the other direction, after roughly 5K steps. However, at this point the preference of each variants against $\pisft$ has already collapsed, indicating a perhaps less meaningful comparison between two sets of bad responses in this region.

To further understand the training dynamics, we plot the log-probabilities of the preferred and dispreferred responses from the preference data for all the variants in Figure~\ref{fig:log_probs}. For the logistic loss, these plots demonstrate that while both the likelihoods rapidly decrease to zero, the rate is faster for the dispreferred responses. In terms of our analysis setup, this corresponds to $\pref^{\pi,\mu}$ being large, which in turn leads to a small value of the curvature constant $c_\base$ and a large value of $\Rscale$. This makes the bound in Theorem~\ref{thm:kl_bound} drastically worse, confirming our theoretical findings empirically. In the case of $\mu = \pisft$, corresponding to DPO, a similar collapse of log probabilities was also observed in prior works~\citep{fisch2024robust, rafailov2024direct}. While the log-likelihoods also decline for the squared loss, the decrease is milder, which in turn means that the magnitude of $\pref^{\pi, \base}$ and the necessary value of $\Rscale$ remain adequately bounded. This, in addition with the better curvature constant of the squared loss makes the findings to be consistent with the theory.

In Table~\ref{tab:c_base} we report approximations to $c_\base$ and $\Rscale$ for all the different loss variants, as measured empirically. We approximate $\Rscale$ by the value of $\omega^{\pi,\base}$ averaged over the last mini-batch of training. Further, we approximate $c_\base$ by the curvature of $\ell_\base$ for the values of $\Rscale$ and $\pref^{\pi, \base}$ obtained this way. We note that $\omega^{\pi,\mu}$ decreases monotonically during training and so the approximation to $\Rscale$ and $c_\base$ is tightest at the end of training. The reported values of $c_\mu$ and $\Rscale$ are consistent with our theory and the empirical observations in Figures\ref{fig:eval} and \ref{fig:log_probs}.
\section{Discussion}

This works adopts a different line of reasoning than the typical reparameterization arguments to motivate the correspondence between offline and online RLHF techniques, with a goal of understanding the impact of design choices in the offline methods, many of which do not fit cleanly in the arguments for equivalence of online and offline methods. Our theory and experiments collectively indicate that perhaps the limited coverage of offline data, and the propensity of log-likelihoods of the preference data to precipitously drop in certain methods are the key obstacles to reliable learning in this setup. Our findings suggest that using losses which do not decay to zero at a slow rate, like the logistic loss, and using experimental design techniques for data collection before offline RLHF can be fruitful avenues for addressing the concerns uncovered here.

\bibliography{pref_theory}
\bibliographystyle{icml2025}

\newpage
\appendix
\onecolumn
\section{From a preference model to a proper loss}
\label{app:preference_to_loss}
A natural family for the reward generating distributions is the exponential family. In particular we are going to assume that $D_\pref$ is in the exponential family parametrized by some $v^\star := v^\star(x,y,y')$, that is $\P(\pref|x,y,y') = \exp(\pref v^\star - \phi(v^\star))$, where $\phi$ is some strictly convex function. Considering the exponential family naturally leads to an objective function for learning the unknown parameter $v$, that is to maximize the log-likelihood which will recover $v^\star$ in the following way.
First we take the derivative of the log-likelihood with respect to $v$
\begin{align*}
    \nabla_v\E_\pref[\log \exp(\pref v - \phi(v))] = \E_\pref[\pref - \nabla \phi (v)] = \nabla \phi(v^\star) - \nabla \phi(v).
\end{align*}
Setting the derivative to $0$ and using the strict convexity to invert $\nabla \phi(v)$ shows that $v^\star = \nabla \phi^{-1}(\E_\pref[\pref])$. To summarize the above, a preference model which follows exponential family distribution gives rise to a natural loss function given by
\begin{align*}
    \min_{v \in [-\Rscale, \Rscale]}E_{\pref}[\phi(v) - \pref v],
\end{align*}
and has a closed form solution $v^\star = \nabla \phi^{-1}(\E_\pref[\pref])$. 

To make this discussion concrete we focus on a BTL model, where we set $v(x,y,y') = R(x,y) - R(x,y')$. Then $\phi(v(x,y,y')) = \log(\exp(R(x,y) - R(x,y')) + \exp(R(x,y') - R(x,y)))$ and the corresponding loss function is then
\begin{align*}
    \phi(v(x,y,y')) - \pref v(x,y,y') &= \log(\exp(R(x,y) - R(x,y')) + \exp(R(x,y') - R(x,y)))\\
    &- \log(\exp(\pref(R(x,y) - R(x,y'))))\\
    &=\log(1 + \exp(-\pref(R(x,y) - R(x,y')))),
\end{align*}
which is precisely the logistic loss. The above derivation already shows that any loss derived from the exponential family with link function $\nabla \phi$ is going to be proper as $\phi$ is strictly convex and hence $\nabla \phi$ is invertable with $g_\ell$ in Assumption~\ref{ass:proper} satisfying $g_\ell \equiv \nabla \phi^{-1}$. We can further check that for a minimizer, $v$, of the logistic loss we must have
\begin{align*}
    \nabla_v \left(\eta \log(1 + \exp(-v)) + (1-\eta)\log(1 + \exp(v))\right) = -\frac{\eta\exp(-v)}{1 + \exp(-v)} + \frac{(1-\eta)\exp(v)}{1 + \exp(v)} = 0,
\end{align*}
where $\eta = \P(\pref = 1|x,y,y')$. The above implies $v = \log\frac{\eta}{1-\eta}$. Equivalently, we could have computed the derivative of the convex conjugate of $\phi$, $\nabla \phi^\star$, which is precisely $\nabla\phi^{-1}$. Finally, to establish the claimed connection between the parametrization of $\Pi$ to the BTL parametrization with $v^\star$ we have the following. The fact that the logistic loss is proper implies
\begin{align*}
    &R(x,y) - R(x,y') - \log\frac{\base(y|x)}{\base(y'|x)} = \pref^{\pi,\mu}(x,y,y')\\
    = &\log\frac{\eta}{1-\eta} = \log \frac{\exp(R^\star(x,y) - R^\star(x,y'))}{ Z(x) - (R^\star(x,y) - R^\star(x,y'))},
\end{align*}
which further simplifying gives Equation~\ref{eq:R-realizable}.

A similar line of reasoning shows that when $\ell(v) = (1 - v)^2$, the link function satisfies $\eta = \frac{1 + v^\star}{2}$ and so the resulting reward model is $\P(\pref|x,y,y') = \frac{1 + R^\star(x,y) - R^\star(x,y')}{2}$.

\section{Experiment details}
\label{app:exp}
We evaluate the different variants on the TL;DR dataset~\citep{volske2017tl}, where the task is to summarize posts on reddit forums. The dataset consists of an original reddit posts, along with a pair of responses which are rated by human judges to provide the groumd-truth preference annotations~\citep{stiennon2020learning}. Our experiments use a T5 large model~\citep{raffel2020exploring} with 770M parameters, which is further fine-tuned to maximize the log-likelihood of the winning responses in the TL;DR dataset.
We train for $20000$ iterations, with a batch size of $32$. A KL regularizer is used to the reference $\pisft$ checkpoint with coefficient equal to $0.005$. The optimizer used is Adafactor with learning rate that is constant with a linear warm-up for $2000$ steps and a base rate of $1e-4$.

We used the following text to prompt the Gemini evaluator used for our experiments:

\texttt{
 You are an expert summary rater who prefers very short and high quality  summaries. Given a document and two candidate summaries, say 1 if SUMMARY1 is the better summary, or 2 if SUMMARY2 is the better summary. Give a short reasoning for your answer.\\
 ARTICLE: \textless article-here \textgreater\\
 SUMMARY1: \textless summary-by-$\pi$\textgreater\\
 SUMMARY2: \textless summary-by-$\piref$\textgreater.
}

\end{document}